\documentclass[10pt,twocolumn,letterpaper]{article}

\usepackage{tpl_3dv/3dv}
\usepackage{times}
\usepackage{epsfig}
\usepackage{graphicx}
\usepackage{amsmath}
\usepackage{amssymb}

\usepackage{mathtools}

\usepackage{tikz}
\usepackage{pgfplots}

\usepackage{subcaption}
\usepackage{enumitem}
\usepackage{xcolor}
\usepackage{style}
\usepackage{style_suppmat}
\usepackage{pict2e}
\usepackage{transparent}
\usepackage{listofitems}
\usepackage{calc}
\usepackage[nomessages]{fp}
\usepackage{booktabs}
\usepackage{arydshln}
\usepackage{comment}
\usepackage{balance}
\usepackage{array, multirow}

\setlist[itemize]{itemsep=0pt,topsep=2pt,partopsep=2pt,parsep=0pt,leftmargin=4mm}
\usepackage[export]{adjustbox}

\usepackage{hyperref}
\hypersetup{pagebackref=true,breaklinks=true,colorlinks,bookmarks=false}

\threedvfinalcopy 

\ifthreedvfinal\fi

\begin{document}

\title{
Representing Shape Collections With Alignment-Aware Linear Models}

\author{
Romain Loiseau\textsuperscript{1, 2}\\
{\tt\footnotesize romain.loiseau@enpc.fr}
\and
Tom Monnier\textsuperscript{1}\\
{\tt\footnotesize tom.monnier@enpc.fr}
\and
Mathieu Aubry\textsuperscript{1}\\
{\tt\footnotesize mathieu.aubry@enpc.fr}
\and
Loïc Landrieu\textsuperscript{2}\\
{\tt\footnotesize loic.landrieu@ign.fr}
\and
{\textsuperscript{1}LIGM, Ecole des Ponts, Univ Gustave Eiffel, CNRS, France}\\
{\textsuperscript{2}LASTIG, Univ. Gustave Eiffel, ENSG, IGN, F-94160 Saint-Mande, France}\\
}

\maketitle
\thispagestyle{empty}

\begin{abstract}
\vspace{-1em}
   In this paper, we revisit the classical representation of 3D point clouds as linear shape models.
   Our key insight is to leverage deep learning to represent a collection of shapes as affine transformations of low-dimensional linear shape models.
   Each linear model is characterized by a shape prototype, a low-dimensional shape basis and two neural networks. The networks take as input a point cloud and predict the coordinates of a shape in the linear basis and the affine transformation which best approximate the input.
   Both linear models and neural networks are learned end-to-end using a single reconstruction loss.
   The main advantage of our approach is that, in contrast to many recent deep approaches which learn feature-based complex shape representations, our model is explicit and every operation occurs in 3D space. 
   As a result, our linear shape models can be easily visualized and annotated, and failure cases can be visually understood. While our main goal is to introduce a compact and interpretable representation of shape collections, we show it leads to state of the art results for few-shot segmentation.
   Code and data are available at: {\small\url{https://romainloiseau.github.io/deep-linear-shapes}}
\end{abstract}
\vspace{-1.5em}
\section{Introduction}\label{sec:introduction}
\begin{figure}[t]
    \centering
    \input{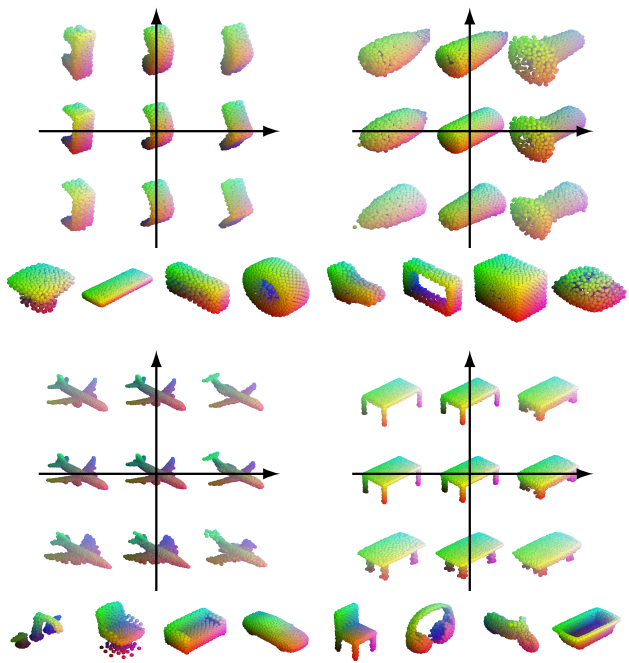}
    \caption{\textbf{Discovered linear models.} Our approach discovers without supervision  linear shape models from large collections of shapes. We show two examples of two-dimensional families and eight additional prototypes discovered for ABC~\cite{koch2019abc} (top) and ShapeNet~\cite{shapenet2015} (bottom).
    }
    \label{fig:teaser}
    \vspace{-1.5em}
\end{figure}
\begin{figure*}[t]
    \centering
    \includegraphics[width=.84\textwidth]{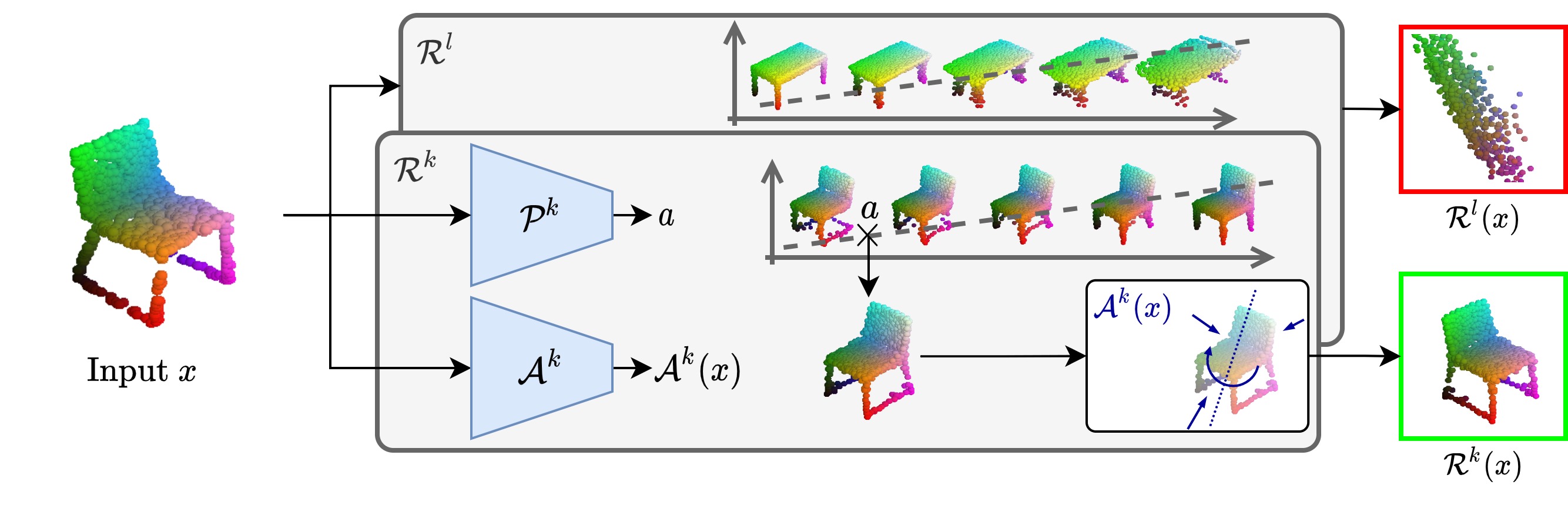}
    \vspace{-1em}
    \caption{\textbf{Method overview.}
    Given an input point cloud $x$, we predict for each shape model $\cR^k$ the element that best reconstructs the input: 
    the projection network $\cPk$ outputs the coordinates $a$ of a shape in a linear family, and the alignment network $\cAk$ predicts the parameters of an affine transformation $\cA^k(x)$ which is applied to the selected shape. The input point cloud is then assigned to the shape model that best reconstructs it, here highlighted in green.}
    \label{fig:pipeline_dti3d}
    \vspace{-1.5em}
\end{figure*}
Picture a company acquiring thousands of 3D scans of technical components; how to leverage, organize, or even simply visualize these 3D models?
Deep shape analysis techniques have flourished over the last years \cite{guo2020deep} but, even when motivated by geometric intuitions, these methods and their results remain hard to interpret and interact with.
Moreover, they are often limited by the availability of domain and application-specific annotations. 
Instead of pushing for even more complex architectures, we
operate directly in 3D space and revisit the simple linear shape model with a deep learning perspective. As illustrated in \figref{fig:teaser}, we model a collection of 3D shapes with a set of low-dimensional linear shape models. Each linear model is defined by a {prototype} 3D point cloud and a set of vector basis that can be interpreted as fields of translation vectors for each point of the prototype. By adding a linear combination of this basis vector to the prototype, one can continuously move in a low-dimensional subspace of the shape space.

We face three key challenges when trying to represent 3D shape collections with such linear models. First, comparing shapes using Chamfer or Earth Mover distances has strong limitations for shape analysis, since they are impacted by simple rigid or affine shape transformations, which cannot be easily represented by linear models. Transformation-invariant distances such as the Gromov-Hausdorff Distance~\cite{memoli2005theoretical} can be defined to overcome this problem, but they are typically very hard to work with. Second, finding the coordinates in the shape basis that best reconstruct a sample according to a given similarity measure is a difficult non-convex problem.
Third, operations as simple as averaging are non-trivial for point clouds, and dimensionality reduction techniques such as Principal Component Analysis~\cite{wold1987principal} do not directly apply.

In this work, we present an unsupervised approach that learns small sets of linear shape models to explain large collections of point clouds. We propose to solve this task with a clustering formulation directly in 3D space, where clusters are associated to linear shape families, each modeled as a reference prototype point cloud and a set of basis vectors that can be interpreted as displacement fields.
We explore two ways of defining such displacement fields - either using a pointwise parametrization or an implicit one based on parametric differentiable functions of 3D space - and analyze their benefits. In addition, to predict the coordinates of a point cloud in the linear basis and account for shape transformations, we extend the idea from the work of Monnier~\etal~\cite{monnier2020dticlustering} on transformation-invariant image clustering to the setting of 3D shape alignment. By jointly learning linear shape families and parametric functions predicting both shape basis coordinates and alignment parameters, our approach is able to discover rich and meaningful shape models from a collection of point clouds without any supervision.

We believe that our method has strong advantages compared to recent unsupervised 3D shape analysis approaches. First, by manipulating objects directly in 3D space, our results are
easy to interpret and visualize.
Second, our linear shape models can serve as a mean to explore large collections of raw 3D point clouds. Finally, we show that despite its simplicity, our model yields competitive results for shape clustering and state-of-the-art results for few-shot shape segmentation.\\
\vspace{-.5em}
\noindent Our contributions can be summarized as follows:
\begin{itemize}
    \item we present an unsupervised method to represent large point cloud collections with a small set of linear families of shapes;
    \item we extend the DTI clustering framework to learn linear shape models by introducing projection networks;
    \item we analyse two different representations for linear shape modeling and show the benefits of representing them with continuous functions of space rather than pointwise displacements;
    \item we demonstrate qualitative results for visualizing the large unstructured ABC dataset~\cite{koch2019abc} and obtain state-of-the-art few-shot segmentation performances on the standard ShapeNetPart dataset~\cite{shapenet2015}.
\end{itemize}
\vspace{-.5em}
\section{Related work}\label{sec:relatedwork}

\paragraph{Point cloud distances and alignment.}
Classical similarity measures between point clouds include the Chamfer and Earth mover distances. These distances are however not invariant to rigid transformations.  The Gromov-Hausdorff distance~\cite{memoli2005theoretical} provides a nice framework to define a transformation-invariance distance, but is difficult to use in practice. 
Transformation-Invariant distances were also defined for images and used for clustering by Frey and Jojic~\cite{freyEstimatingMixtureModels1999, freyFastLargescaleTransformationinvariant2002,  freyTransformationinvariantClusteringUsing2003,tcafreyjojic}.  
The Transformed Component Analysis (TCA) approach ~\cite{tcafreyjojic} would be particularly relevant, although operating with a discrete set of transformations may be too limiting for aligning 3D shapes. Applying them to 3D point clouds would require to align them. 
This is classicaly done using the the Iterative Closest Point (ICP) algorithm ~\cite{besl1992method}. Instead, we take inspiration from the Deep Transformation-Invariant framework~\cite{monnier2020dticlustering} and use neural networks to predict alignment and define similarity.

\vspace{-1em}
\paragraph{Linear Shape Modeling.} The idea of representing a collection of images using a low-dimensional image basis was first developed for face images~\cite{sirovich1987low}. Popularized by the classical eigenfaces model~\cite{ turk1991eigenfaces}, linear models have since been applied to diverse computer vision problems and data. A linear 3D face model was designed in~\cite{blanz1999morphable} and applied to new view synthesis. \cite{cootes2001statistical} demonstrated applications to medical data.
Non-rigid surface-from-motion can also benefit from linear shape basis decomposition to recover 3D shapes~\cite{bregler2000recovering, torresani2008nonrigid, dai2014simple}. An application of linear modeling to unsupervised 3D keypoint discovery was recently demonstrated in~\cite{fernandez2020unsupervised}.
These linear models are typically learned from a set of examples by principal component analysis, factorization techniques \cite{tomasi1992shape}, or defined manually \cite{wang2015linear}. Additionally, some recent works propose to analyse shape collections through implicit representations~\cite{jiang2020shapeflow, zheng2021deep, deng2021deformed}. In contrast, we propose a learning-based approach to model arbitrary unregistered shapes from large collections of examples, and we use several low-dimensional linear families.

\vspace{-1em}
\paragraph{Deep Learning for 3D Analysis.} 
Neural networks successfully tackled numerous challenges in 3D shape analysis. The main approaches can be broadly classified depending on the representation of 3D data they leverage,  voxels~\cite{choy20194d}, point clouds~\cite{qi2017pointnet}, graphs \cite{landrieu2018large}, surfaces~\cite{bronstein2017geometric,groueix2018papier}, structured models~\cite{li2017grass,genova2019learning}, or more recently implicit volumetric models~\cite{park2019deepsdf,chen2019learning,mescheder2019occupancy}. They have been successfully applied to tasks as diverse as classification~\cite{manessi2020dynamic}, segmentation~\cite{thomas2019kpconv}, shape generation~\cite{groueix2018papier,tatarchenko2019single}, matching~\cite{pais20203dregnet}, denoising~\cite{hermosilla2019total} and compression~\cite{huang2020octsqueeze}. In this paper, we focus on 3D point clouds but our approach is general and could be extended to other representations.

The key idea to use Multi-Layer Perceptron (MLP) on point clouds was initially proposed for shape classification and segmentation by Qi \etal~\cite{qi2017pointnet} with an architecture called PointNet, and for 3D point cloud generation in~\cite{fan2017point}. To answer the difficulty of annotating 3D data, new approaches are able to perform self-supervised and unsupervised feature learning \cite{yang2018foldingnet, sauder2019self, hassani2019unsupervised} and low-shot segmentation \cite{groueix2019unsupervised, wang2020few, gadelha2020label}. Especially related to ours is the recent 3D capsule approach~\cite{zhao20193d} that explicitly tries to design a shape representation invariant to 3D transformations and can be applied to many tasks. However, the latent representations learned with 3D capsules and the associated generation process are difficult to interpret.
Also related to ours is the approach of Deprelle~\etal~\cite{deprelle2019learning} which proposes a 3D shape reconstruction model obtained by combining and transforming learned elementary structures. This method shares similarities with ours as it allows us to learn prototypes of parts of shapes. However, it focuses on reconstruction accuracy, uses a single prototype per part and mainly follows the black-box AtlasNet~\cite{groueix2018papier} deformation framework.
\vspace{-.5em}
\section{Modeling Shape Collections}

\label{sec:method}
Our goal is to explain a collection of $N$ point clouds $\xn$ with a small set of $K$ shape models. For simplicity, we assume that all point clouds have the same number $M$ of points. We propose to solve this task with a clustering formulation described in \Secref{sec:clustering}. We then describe how we model alignment (\Secref{sec:align_aware}) and linear shape families (\Secref{sec:linearmodels}) resulting in our final modeling. Finally, we present how we parametrize our linear shape models and give some training details (\Secref{sec:param}).

\subsection{Method overview}
\label{sec:clustering}
We build a set of $K$ shape models $\cR=\{\cR^1,\ldots,\cR^K\}$.
Each $\cRk$ maps a sample point cloud $x$ to a reconstructed point cloud $\cRk(x)$ which can be interpreted as the approximation of $x$ by the corresponding model. We denote by $d$ a distance between point clouds which measures the quality of a reconstruction. We use the Chamfer distance in all our experiments.
We learn the shape models $\cR$ by minimizing the loss
\begin{align}
  \cL(\cR)
&=\sum_{x\in \xn} \min_{k=1}^K  
d
\left(x,\cRk(x)\right)~,
\label{eq:loss}
\end{align}
which can be interpreted as a clustering objective defined as the sum of the reconstruction errors with optimal cluster assignment.

\vspace{-1em}
\paragraph{Prototype model.}
The simplest form of $\cRk$ is a constant function:  $\cRk_\text{proto}(x)=\ck\in\bR^{M \times 3}$ where each $\ck$ can be seen as a prototype point cloud. Such prototype point clouds can be learned by minimizing $\cL$ with batch Stochastic Gradient Descent (SGD). This 
amounts to performing stochastic K-means~\cite{bottou1995convergence} for $3$D point clouds. Note that this is a weak reconstruction model, however, the goal of this paper is not to learn the most faithful reconstruction, but rather to summarize the collection.
\subsection{Alignment-Aware Model}\label{sec:align_aware}
A clear limitation of the prototype model is that it does not take into account simple geometric transformations of the point clouds, such as rigid transformations.
For example, point clouds can be close to a model's prototype $\ck$ according to the distance $d$, while a rotated or translated version of the same point cloud is far away.
We would like both point clouds to be associated with the same shape model. 
To address this issue, we incorporate in each model $\cRk$ an affine alignment component. In practice, we use neural networks $\cAk$ - which we refer to as \emph{alignment networks} - whose goal is to predict an affine transformation $\cAk(x)$ 
aligning the prototype $\ck$ with a target point cloud $x$. This results in an alignment-aware model $\cRk_\text{align}$ defined by:
\begin{equation}
    \cRk_\text{align}(x)=\cAk(x)\left[c^{k}\right]~,
    \label{eq:Rdti}
\end{equation}
\noindent where the affine transformation $\cAk(x)$ is applied to each point of the prototype point cloud $\ck$. The alignment networks $\cA^{1},\ldots,\cA^{K}$ can be trained alongside the prototypes $c^{1},\ldots,c^{ K}$ by minimizing \eqref{eq:loss}. This model can be seen as an extension of the recent Deep Transformation-Invariant (DTI) clustering framework~\cite{monnier2020dticlustering} developed for images to point clouds. Indeed, our alignment models can be understood as defining an approximation of an affine-invariant version of the distance $d$ according to which the clustering is performed. In this paper, we rather view these networks as an integral part of the shape models. 

Note that different transformation models could be considered. In our experimental analysis, we study variations of the model using weaker transformations, such as rigid transformations or scaling, and show the benefits of the affine model. On the contrary, one could consider complex deformations parametrized by deep networks, such as the ones used in FoldingNet~\cite{yang2018foldingnet} or AtlasNet~\cite{groueix2018papier}, which would surely lead to higher accuracy reconstructions. However, such transformations completely change the geometry of a point cloud and are hard to interpret.
\subsection{Linear Shape Modeling}
\label{sec:linearmodels}
Our goal in this section is to model changes in objects more subtle than those that can be modeled by affine transformations, such as the angle of the wings of an airplane, while maintaining the model intepretability. We propose to associate a linear shape family to each prototype point cloud.
\vspace{-1em}
\paragraph{Linear shape families.}
\label{sec:shape_model}
For each model $k$, we define a linear shape family as a pair formed by 
(i) a prototype point cloud $\ck$ in $\bR^{M \times 3}$ and  
(ii) a set $\vk$ of $D$ basis vectors $\vk=\{\vk_1,\ldots,\vk_D\}$, where each $\vk_i \in \bR^{M \times 3}$ associates to each point of the prototype a 3D vector and can be interpreted as displacement fields. Each $(\ck,\vk)$ defines a continuous collection of shapes covered by translating the points of $\ck$ along the directions defined by $\vk$. Each element $u$ of the linear family $(\ck, \vk)$ is characterized by a vector $a$ in $\bR^D$ defining its coordinates in the linear shape family:
\begin{equation}
    u = \ck
    +  \sum_{i=1}^D a_i\,\vk_{i}~.
    \label{eq:linear}
\end{equation}
\noindent 
The vector $a$ can be interpreted as the set of amplitudes to apply to the displacement fields $\{\vk_1,\ldots,\vk_D\}$. Note this formally describes an affine space but we follow the convention of previous works and refer to it as linear.\footnote{An analogy can be made with the face reconstruction model EigenFace~\cite{turk1991face}: $c$ is equivalent to the \emph{mean face}, and $v$ to the eigenfaces.} Also note that we do not explicitly enforce linear independence between basis vectors, but their high dimensionality ($M\times 3$) leads to such independence in practice.
\vspace{-1em}
\paragraph{Projection networks.} If we had access to ordered point clouds, \ie lists of $M$ points in $\bR^3$ where the $i$-th points are in correspondence, we would be able to use the $L_2$ distance to measure point clouds similarity. In this case, computing the coordinates of the element of the linear family closest to a target point cloud would simply amount to performing Euclidean projection. This is however not the case for unordered point clouds, for which the notion of distance is more complicated. For common point cloud similarity measures such as the Chamfer distance, finding the closest point cloud in a linear family is a difficult non-convex optimization problem. This task is made even harder by the fact that we 
use our alignment networks to transform the elements of the family before comparing them with the input cloud.

Therefore, we propose to leverage deep learning to estimate which element of a linear family is the closest to a target point cloud after alignment. More specifically, we associate to each linear family $(\ck,\vk)$ a neural network $\cPk$ which aims at associating to a given input sample the coordinates of the element in the linear family minimizing the distance $d$. The output of the network $\cPk(x)\in\bR^D$ is interpreted as the coordinates $a$ of the point cloud defined in \eqref{eq:linear}. By analogy with the $L_2$ distance case, we refer to these networks as \emph{projection networks}.

\vspace{-1em}
\paragraph{Full model.} We define our final shape model $\cR$ as a collection of models $\cRk_\text{full}$ each composed of a linear family $(\ck,\vk)$, an alignment network $\cAk$ and a projection network $\cPk$.
Given a target point cloud $x$, our model reconstructs it by
(i) selecting an element of the linear family $(\ck,\vk)$ through the projection network $\cPk$, and 
(ii) aligning it with the target using the transformation predicted by the alignment network $\cAk$. More formally, we write each shape model as:
\begin{equation}
    \cRk_\text{full}(x)=\cAk(x) \left[c^{k}+ \sum_{i=1}^D \left[\cPk(x)\right]_i \vk_i\right],
    \label{eq:reconstruction}
\end{equation}
\noindent where $\left[\cPk(x)\right]_i$ refers to the i-th component of $\cPk(x)$ and the affine transformation $\cAk(x)$ is applied to each point of the point cloud independently. Again, we optimize jointly the $\ck, \vk, \cAk$ and $\cPk$ to minimize the reconstruction loss defined in Equation~(\ref{eq:loss}).

\begin{figure*}[t]
    \centering
    \input{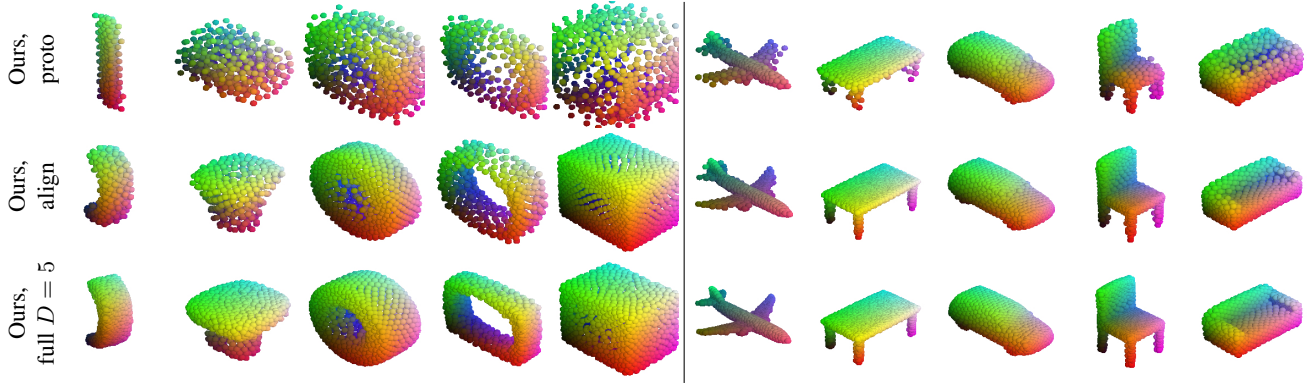}
    \caption{\textbf{Learned prototypes and comparisons.} We compare the prototypes from our different shape modeling discovered in ABC~\cite{koch2019abc} (left, $5$ shape models out of $10$) and ShapeNetCore~\cite{shapenet2015} (right, $5$ shape models out of $55$). Note how sharp the prototypes become when the shape modeling complexity increases, respectively with alignment-awareness and 5-dimensional linear families.
    }
    \label{fig:qualitative_abc_all_clusters}
    \vspace{-1.5em}
\end{figure*}

\subsection{Parameterization and training details}\label{sec:param}

We first describe how we parametrize the linear families, then provide implementation details such as networks architecture and our curriculum learning strategy.

\vspace{-1em}
\paragraph{Linear family parametrization.}
While the prototypical point cloud  $\ck$ is modeled directly using learnable parameters in $\mathbb{R}^{M\times 3}$, the basis vectors $\vk_i$ can be parametrized in two different ways:
\begin{itemize}
\item \emph{Pointwise parametrization:} for each model $k$, we represent $\vk$ as vectors of learnable parameters of size $D\times(M \times 3)$ that can directly be interpreted as $D$ pointwise displacement vectors of the prototype $\ck$.
\item \emph{Implicit parametrization:} we use implicit parametric functions of the 3D space modeled as neural networks to define the displacement fields. More precisely, for each model $k$ and basis dimension $i$, we learn a parametric function $\mathcal{V}_i^k:\bR^3 \mapsto \bR^3$ mapping any point in the 3D space to a displacement direction.
Writting $[\ck]_p$ the 3D coordinates of the $p$-th point of prototype $\ck$, the 3D coordinates $[\vk_i]_p$ of the $i$-th basis vector associated to the point $p$ are  $[\vk_i]_p=\mathcal{V}_i^k([\ck]_p)$.
\end{itemize}
\noindent Intuitively, the pointwise parametrization seems better suited for modeling complex and discontinuous transformations within a shape family such as the appearance/disappearance of object parts. On the contrary, the transformations learned with implicit parametrizations are derived from continuous functions of the 3D space and can be expected to be more regular. 

We compare both settings in \secref{sec:quant}, and show that pointwise parametrizations provide better shape reconstructions, but that implicit parametrization yields more interpretable transformations preserving semantic correspondences. Thus, unless specified otherwise, we use the implicit parametrization of the basis in the rest of the paper.

\vspace{-1em}
\paragraph{Architecture.} For each model $k$, the alignment network $\cAk$ takes as input a point cloud and outputs a vector in $\bR^{12}$ corresponding to a linear 3D operator and a translation vector applied to each point of the model. The projection network $\cPk$ also takes a point cloud as input and outputs a vector in $\bR^{D}$ that is interpreted as coordinates in the linear family $(\ck,\vk)$. These networks share a common PointNet~\cite{qi2017pointnet} backbone encoder which acts as a global feature extractor. This shared encoder starts with a sequence of three linear layers with batch normalization~\cite{ioffe2015batch} and ReLU activation acting on points independently and sequentially generating representations of size 64, 128 and 1024, and ends with a max-pooling over all points.
This encoder is then followed by $2\times K$ Multi-Layer Perceptrons (MLPs) 
corresponding to each prediction task (alignment or projection) and each shape model. Each MLP has one hidden layer of size 128. The implicit parametrizations $\cV^k_i:\bR^3 \mapsto \bR^3$ are MLPs with 2 hidden layers of size 128.

\vspace{-1em}
\paragraph{Curriculum learning.} Inspired by the curriculum learning of~\cite{monnier2020dticlustering}, we propose to learn our models by gradually increasing the models complexity. We first learn raw prototype models ($\cRk_\text{proto}$), an optimization which corresponds to performing a gradient-based K-means algorithm in the 3D space. Second, we augment each model with alignment awareness ($\cRk_\text{align}$). Finally, we gradually increase the linear families dimension up to the desired one, resulting in our final shape model ($\cRk_\text{full}$).

\vspace{-1em}
\paragraph{Implementation details.}
Our implementation - which will be released upon publication - uses PyTorch, Torch-Points3D \cite{tp3d}, and an efficient CUDA implementation of the Chamfer distance which significantly speeds up training.
With $K=10$ prototypes and $D=5$, our model has $4.6M$ parameters. For comparison, the reconstruction models proposed by Wang \etal \cite{wang2020few} and Groueix \etal \cite{groueix2019unsupervised} have respectively $2.6M$ and $10.0M$ parameters.
See our supplementary material for additional details.
\vspace{-.5em}
\section{Experiments}
In this section, we analyze the benefits of our method to represent shape collections, first qualitatively (Section~\ref{sec:qual}) then quantitatively  (Section~\ref{sec:quant}).
Finally, we demonstrate that it leads to results on par with state of the art for few-shot and low shot  shape segmentation (Section~\ref{sec:segm}).
\subsection{Qualitative results}\label{sec:qual}

\begin{figure}[t]
    \centering
    \input{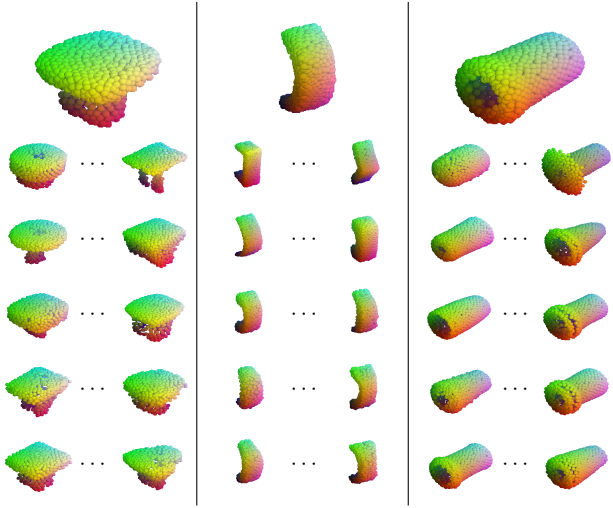}
    \caption{
    {\textbf{Basis vectors.}} Examples of linear shape models obtained after training on ABC with $D=5$. The prototype is represented at the top and each row corresponds to one of the dimension of the linear families. The models' basis vectors correspond to complex morphological changes.
    }
    \vspace{-1.5em}
    \label{fig:linear}
\end{figure}

We demonstrate the potential of our method for exploring large shape 
collections.

\vspace{-1em}
\paragraph{Datasets.}
The ShapeNet dataset~\cite{shapenet2015} is a large collection of over $50K$ 3D models organized along $55$ common object categories such as chairs, airplanes, or cars. The ABC dataset~\cite{koch2019abc} is a very large collection of Computer-Aided Design (CAD) models of diverse mechanical object parts, such as screws or pipes. We used the first six chunks from this dataset and considered the connected components of each mesh as separate objects ($\approx70K$ shapes). We apply our approach using 55 shape models for ShapeNet and 10 for ABC. For both datasets, we uniformly sample points on the objects’ surface to obtain point clouds.
\vspace{-1em}
\paragraph{Prototypes.}We present in Figure \ref{fig:qualitative_abc_all_clusters} examples of prototypes learned when successively adding different components of our method.
The first line, denoted {``Ours, proto"}, represents the linear families' prototypes learned during the first stage of our training ($\cR_\text{proto}$).
The second line, denoted ``Ours, align", displays the learned prototypes after the second stage of our training ($\cR_\text{align}$), during which affine alignment networks are learned jointly with their model's prototype.
Finally, the third line denoted ``Ours, full $D=5$" illustrates the prototypes learned at the last stage of training ($\cR_\text{full}$) alongside linear shape families of dimension $5$ and their associated projection networks. We show the center of each linear shape model, defined by taking the median amplitude $a_i$ in each dimension $i$ when considering all point clouds associated with the model, \ie point clouds for which this model outputs the best reconstruction.

The prototypes learned with the Chamfer distance (first line) appear noisy, hinting that they are not well aligned with the shapes they try to approximate. When adding alignment networks, we obtain the prototypes of the second line, which are much cleaner, outlining the interest of using a transformation-invariant model, as well as the fact that our approach can effectively learn such a model. Finally, the prototypes obtained with our full method are even sharper and smoother, indicating that linear shape families can better model the associated point clouds.

Our results on the ABC dataset outline the capacity of our full model to differentiate between different types of shapes, as prototypes correspond to different object types.
By looking at the prototypes, one can grasp at a glance the diversity of shapes contained in this large-scale dataset.
\vspace{-1em}
\paragraph{Linear shape models.}
In~\figref{fig:linear}, we illustrate some of the linear shape models learned on ABC (more results for both datasets are in the supplementary material). The top row shows the center of the linear shape models, and the subsequent lines illustrate the five basis vectors.
For each model and each basis vector, we represent two shapes whose amplitudes for the considered dimensions are set to the $5$-th and $95$-th percentile values of all point clouds associated to the model, while the other amplitudes remain at the median value. Again, we can see how the different dimensions give insights on the
diversity of shapes within the dataset.

\vspace{-1em}
\paragraph{Reconstructions.}In \figref{fig:unsup_annotated_clusters}, we show examples of reconstructed shapes from ShapeNet (airplanes, cars, and chairs) for four different linear shape families. As expected, the model is able to reconstruct objects precisely while remaining visually interpretable. Again, more examples can be seen in supplementary material.

\subsection{Quantitative Analysis for Clustering and Reconstruction}\label{sec:quant}

\begin{table}[t]
\caption{\textbf{Results on ModelNet10.} We present results with 10 linear shapes models, first for different restrictions of the alignment networks, then for different basis vector configurations. The  steps in the curriculum training of our model are in {\bf bold}. We report clustering accuracy in \% ('Accuracy') and the Chamfer distance multiplied by $10^3$ ('CD'), Results are averaged over five runs.}
    \vspace{-1em}
    \small
    \centering
    \begin{tabular}{@{}lll|cc@{}}
     \toprule
         & & & Accuracy & \multicolumn{1}{c}{CD} \\\midrule
          & \multicolumn{2}{l|}{\textbf{Ours, proto}}  & $\mathbf{63.9\pm1.5}$   & $\mathbf{20.0\pm0.4}$  \\
          &  \multicolumn{2}{l|}{\textcolor{gray}{... with supervision }}& \textcolor{gray}{$79.0\pm0.2$} & \textcolor{gray}{$23.5\pm0.0$}
        \\\midrule 
        \parbox[t]{1.5mm}{\multirow{6}{*}{\rotatebox[origin=c]{90}{Ours, align}}}
        & \multicolumn{2}{l|}{Rigid transformation {(6D)}} &   $64.6\pm5.2$  &    $16.2\pm0.1$   \\
        & \multicolumn{2}{l|}{Trans. + Iso. Scaling (4D)}  &  $71.5\pm4.1$   &   $15.0\pm0.1$    \\
        & \multicolumn{2}{l|}{Trans. + Aniso. Scaling (6D)}  &  $74.1\pm3.0$   &  $10.4\pm0.1$     \\
        & \multicolumn{2}{l|}{Linear (9D)} &  $71.85\pm4.7$   &    $11.1\pm0.1$   \\
        & \multicolumn{2}{l|}{\textbf{Affine (12D)}}    & $\mathbf{75.9\pm3.0}$   & $\mathbf{9.7\pm0.0}$  \\
        &  \multicolumn{2}{l|}{\textcolor{gray}{... with supervision }}& \textcolor{gray}{$88.9\pm0.5$}   &\textcolor{gray}{$11.2\pm0.0$}
        \\\midrule
        \parbox[t]{1.5mm}{\multirow{5}{*}{\rotatebox[origin=c]{90}{Ours, full}}}
        & \parbox[t]{1.5mm}{\multirow{2}{*}{\rotatebox[origin=c]{90}{$D=1$}}} & Pointwise parametrization & $74.3\pm1.7$ & $7.9\pm0.0$\\
        & & \textbf{Implicit parametrization}    & $\mathbf{77.5\pm2.8}$   & $\mathbf{8.1\pm0.0}$  \\
        & & \textcolor{gray}{... with supervision }& \textcolor{gray}{$89.7\pm0.6$}   &\textcolor{gray}{$9.5\pm0.0$}\\
        \cdashlinelr{2-5}
        & \parbox[t]{1.5mm}{\multirow{3}{*}{\rotatebox[origin=c]{90}{$D=5$}}} & Pointwise parametrization & $75.1\pm1.7$ & $5.7\pm0.0$\\
        & & \textbf{Implicit parametrization}   & $\mathbf{77.0\pm3.4}$   & $\mathbf{5.9\pm0.0}$  \\
        & & \textcolor{gray}{... with supervision }& \textcolor{gray}{$90.4\pm1.0$}   &\textcolor{gray}{$7.8\pm0.0$}\\\midrule
        \multicolumn{3}{l|}{FoldingNet~\cite{yang2018foldingnet}} & $76.3\pm7.5$ & $\mathbf{3.5\pm0.0}$ \\
    \bottomrule
    \end{tabular}
    \vspace{-1.5em}
    \label{tab:clu_MN10_10clu}
\end{table}
The qualitative results described in the previous section outline the potential of our approach for visualizing and analyzing large, unstructured, and diverse shape collections. We now provide a more quantitative analysis of these results on the standard ModelNet10 dataset~\cite{wu20153d}. 
\vspace{-1em}
\paragraph{Data and evaluation.} ModelNet10 contains $3991$ train and $909$ test aligned 3D point clouds obtained from CAD models of 
$10$ different classes. 
We use this dataset both in its original aligned version and also with added random rotations around the z-axis to evaluate the capacity of our method to represent unaligned data. Unless specified otherwise, the results are given for the original dataset.
We trained the different variants of our method with $10$ reconstruction models on train and test shapes of ModelNet10. We evaluate in \tabref{tab:clu_MN10_10clu} the clustering accuracy and reconstruction error measured by the Chamfer Distance. To measure the quality of the resulting clustering, we assign to each model the majority label of its associated point clouds from the train set. 
The accuracy of the classification is then defined by assigning to test shapes the label of the model giving the best reconstruction.
\vspace{-1em}
\paragraph{Alignment.} 
We compute the performance of our models only defined by prototypes (``Ours proto"), and then train models with alignments of different complexities (``Ours, align"). 
We first evaluate a model whose alignment networks are restricted to a rigid transformation (``Rigid transformation (6D)"), with rotations parametrized with quaternions. We also evaluate models with a scaling and a translation (``Trans. + Iso. Scaling (4D)"), axis-aligned scalings and a translation (``Trans + Aniso. Scaling (6D)"), a linear transformation (``Linear (9D)"), and finally an affine transformation (``Affine (12D)").
We observe that using alignment networks allows significant clustering improvement in terms of accuracy and reconstruction quality. Moreover, restricting the output of the alignment networks leads to a lower performance: even for centered and rotation-aligned data such as ModelNet, allowing complex alignments benefits both clustering and reconstruction.

\vspace{-1em}
\paragraph{Linear families.} We then evaluate models with affine alignment but different linear basis (``ours, full"). We compare the results between one-dimensional ($D=1$) and five-dimensional ($D=5$) linear families as well as between basis vectors learned in the pointwise and implicit parametrization (see \secref{sec:shape_model}).
Increasing the dimension of the shape families improves the reconstruction error but slightly decreases the clustering accuracy with the implicit parametrization.
This can be explained by the models becoming too expressive, resulting in point clouds from different classes being associated with the same model.
\vspace{-1em}
\paragraph{Baseline and supervised upper bound.} 
As a baseline, we performed k-means clustering in feature space using the implementation of  FoldingNet~\cite{yang2018foldingnet} proposed by~\cite{tao2020}.
The resulting accuracy is comparable to that of our best models'. However, FoldingNet relies on learning black-box deep deformations of a planar patch, and the resulting shape family and generation process are thus 
harder to interpret than ours.

We also trained our model in a supervized manner by associating a class to each model, and only training each model on point clouds from their class (``with supervision" lines, in light gray).
As expected, this "oracle" setting performs better in terms of clustering accuracy, but with lower reconstruction quality. 
This can be explained by the presence of classes with high variability such as \emph{chairs} which require several families to fully cover, and similar classes such as \emph{desks} and \emph{tables} which can be well reconstructed by a single family.

\begin{table}[t]
    \caption{\textbf{Non-aligned data.} Clustering Accuracy ('Accuracy', in \%) and reconstruction error ('CD', Chamfer distance multiplied by $10^3$) obtained with 10 linear shapes models on the rotated version of ModelNet10. $\Delta_{\text{CD}}$ is the difference of reconstruction error when training the same model on the aligned or unaligned datasets.
    }
    \vspace{-1em}
    \small
    \centering
    \addtolength{\tabcolsep}{-2pt}
    \begin{tabular}{@{}l|cccc@{}}
        \toprule
         & Accuracy & $\Delta_{\text{Accuracy}}$ & CD & $\Delta_{\text{CD}}$ \\\midrule
        Ours, proto & $41.2\pm3.4$ &   $-22.7$      & $30.1\pm0.1$ & $-10.1$\\
        Ours, align & $61.8\pm3.3$ &    $-14.1$     & $11.0\pm0.1$ & $-1.3$\\
        Ours, full $D=1$ & $65.2\pm6.7$ &    $-12.3$     & $9.3\pm0.0$ & $-1.2$\\
        Ours, full $D=5$ & $\mathbf{68.8\pm7.9}$ &     $\mathbf{-8.2}$    & $\mathbf{6.7\pm0.0}$ & $\mathbf{-0.8}$\\
        \bottomrule
    \end{tabular}
    \label{tab:unaligned_data}
    \vspace{-1.75em}
\end{table}

\vspace{-1.1em}
\paragraph{Non-aligned data.} In \tabref{tab:unaligned_data}, we report our approach's performance when trained on ModelNet10 with random rotations. We observe that adding alignment networks to the model results in significantly better metrics compared to simple prototypes. Our full models with alignment are able to reach reconstruction qualities almost comparable to the equivalent models trained on aligned shapes. Similarly, the drop in clustering performance is reduced when adding the alignment networks and linear shape families. This outlines the capacity of our models to handle raw unaligned data. We present in the appendix illustrations of the prototypes learned in this setting.
%
\vspace{-.5em}
\subsection{Application to few/low-shot Segmentation}\label{sec:segm}

\begin{table*}[t]
    \caption{\textbf{$\mathbf{10}$-shot segmentation.} We report pointwise IoU for $9$ classes and the average IoU over all $16$ classes of ShapeNetPart. See text for details.
    }
    \vspace{-1em}
    \centering
    \small
    \begin{tabular}{@{}clcccccccccc@{}}
    \toprule
        & & airplane & bag & cap & car & chair & lamp & laptop & mug & table & avg \\\midrule
        \multirow{2}{*}{\textbf{Shared}} & Gadelha \etal~2020~\cite{gadelha2020label}
        & --- & --- & ---  & --- &--- & --- &--- &--- &---&$74.1$\\
        \multirow{2}{*}{\textbf{encoder}} & Ours, full $D=5$ (random)
        & $71.7$    & $70.6$ & $\mathbf{84.0}$  & $62.1$  & $78.8$  & $68.7$  &  $93.1$  &  $87.5$  & $70.6$ & $72.5$ \\
        & Ours, full $D=5$ (prototype) &  $\mathbf{79.4}$   & $\mathbf{73.0}$ & $81.8$ & $\mathbf{72.1}$  & $\mathbf{83.6}$  & $\mathbf{76.1}$   & $\mathbf{94.7}$ & $\mathbf{89.8}$ & $\mathbf{76.2}$  &  $\mathbf{77.4}$\\
        \midrule
        & Wang \etal~2020~\cite{wang2020few} & $67.3$  & $74.4$ & $\mathbf{86.3}$  & --- &$83.4$ &$68.7$  & $93.8$ & $90.9$ &$74.2$&---\\
        \textbf{One encoder} & Groueix \etal~2019~\cite{groueix2019unsupervised} & $67.1$ & --- & ---  & $61.4$& $78.9$& $65.8$ & --- &--- & $66.1$&---\\
        \textbf{per class} & Ours, full $D=5$ (random)  & $72.2$    & $66.0$ & $75.5$  & $63.0$  & $79.1$ & $68.9$  & $93.1$ & $84.2$ & $69.4$ & --- \\
        & Ours, full $D=5$ (prototype) &  $\mathbf{80.0}$ & $\mathbf{79.7}$ & $76.1$  & $\mathbf{72.0}$  & $\mathbf{83.6}$  & $\mathbf{77.1}$   & $\mathbf{94.9}$ & $\mathbf{91.1}$ &  $\mathbf{75.9}$ &  ---\\
    \bottomrule    
    \end{tabular}
    \label{tab:fewshot_segmentation}
    \vspace{-1em}
\end{table*}

\begin{figure}[t]
    \centering
    \input{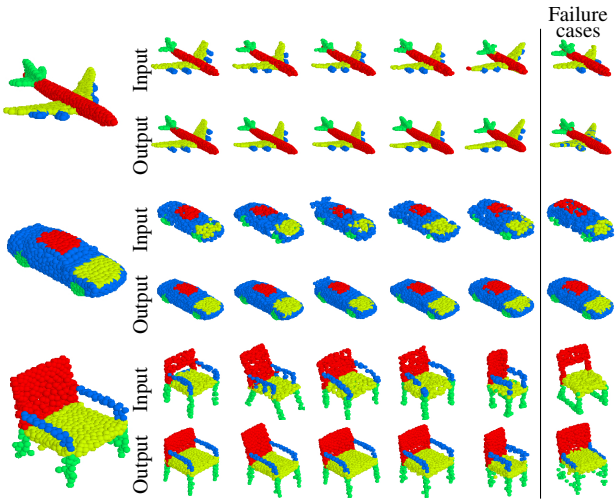}
    \caption{\textbf{Reconstruction results.} Examples of samples annotated pointwise with our semantic-segmentation method ($D=5$). We visually selected failure cases where semantic regions were wrongly predicted. {\small Prototypes are represented on the left column, the ``input" lines display input samples with ground truth annotations and the ``output" lines our reconstruction with pixel labels propagated from the prototype. }}
    \label{fig:unsup_annotated_clusters}
    \vspace{-2em}
\end{figure}
\label{sec:fewshot}

Our linear shape models can perform semantic segmentation by transferring point labels from the model's prototype to the reconstructed point cloud. More precisely, given an input point cloud $x$, we identify the model $k$ with the lowest reconstruction error. We then compute $\tilde{x}=\cRk_\text{full}(x)$, the point cloud reconstructed by this model. 
We transfer the point annotation from the prototype $\ck$ to $\tilde{x}$.
Finally, each point of $x$ is assigned the label of the closest point of $\tilde{x}$. This strategy is especially meaningful in a few-shot setting, since only the prototypes need to be annotated.

\vspace{-2em}
\paragraph{{Few-shot Segmentation.}}In this setting, where we only use a few annotations for each class and train our model with only the reconstruction loss as described earlier, we consider two methods to annotate the prototypes: 
\begin{itemize}
    \item {\it Random}. We randomly pick one sample from the train set for each model and propagate its labels to their nearest points of the aligned prototype.
    \item {\it Prototype.} We align all samples from the train set for each model's prototype and label each point with majority voting. This second setting is meant to emulate the \emph{manual} annotation of the $10$ prototypes. While this is not directly comparable to other approaches, it outlines the crucial advantage given by our approach, which identifies a small set of prototype shapes that can be annotated instead of using random samples. Some prototypes annotated in this manner can be seen in \figref{fig:unsup_annotated_clusters}.
\end{itemize} 

\noindent 
We use the densely annotated ShapeNetPart~\cite{savva2015semantically} to evaluate the segmentation performance of our few-shot segmentation scheme. We report in \tabref{tab:fewshot_segmentation} the performance of our $10$-shot segmentation scheme for nine classes of ShapeNetCore, and the average performance over all 16 classes. 
{As mentioned in \secref{sec:param}, all the alignment and projection networks share a common PointNet~\cite{qi2017pointnet} encoder which acts as a global
feature extractor. To compare with previous works that use either a shared model or a different model per class, we present results using either a single encoder for all classes or one encoder per class. } 
Using only $10$ samples from the dataset to annotate our prototypes,
we observe that the annotation from random samples performs on par or better than state-of-the-art approaches.
Annotating prototypes (using all training samples) significantly outperforms all methods. This shows that our approach can be used to precisely and densely annotate large shape datasets with minimal human intervention.
We also observe some failure cases shown in the last column of Figure~\ref{fig:unsup_annotated_clusters}: since our model can only move points and not add or subtract them, shapes with optional parts, such as the arms of chairs, may be mislabeled.

\vspace{-1em}
\paragraph{{Low-shot Segmentation.}}
Our model can also be trained in a low-shot setting, yielding a slight improvement of +1 and +2 mIoU compared to 3D capsules~\cite{zhao20193d} when trained with only $1$\% or $5$\% of annotated shapes. More details on these results are provided in the supplementary material.
\vspace{-.5em}
\section{Conclusion}
\vspace{-.5em}
We presented a new take on linear shape models with deep learning, representing large un-annotated collections of 3D shapes. Our alignment-aware model produces concise, expressive and interpretable overviews of unaligned point clouds collections. We 
show that our method 
leads to state-of-the-art results for few-shot segmentation.

{
\vspace{-1em}
\paragraph*{Acknowledgements}{This work was supported in part by ANR project READY3D ANR-19-CE23-0007 and HPC resources from GENCI-IDRIS (Grant 2020-AD011012096). We thank François Darmon, Damien Robert, Vivien Sainte Fare Garnot and Yang Xiao for inspiring discussions and valuable feedback.
}}

\clearpage

\nobalance{
\part*{Supplementary material}


\section{Implementation details}

\begin{figure*}[ht!]
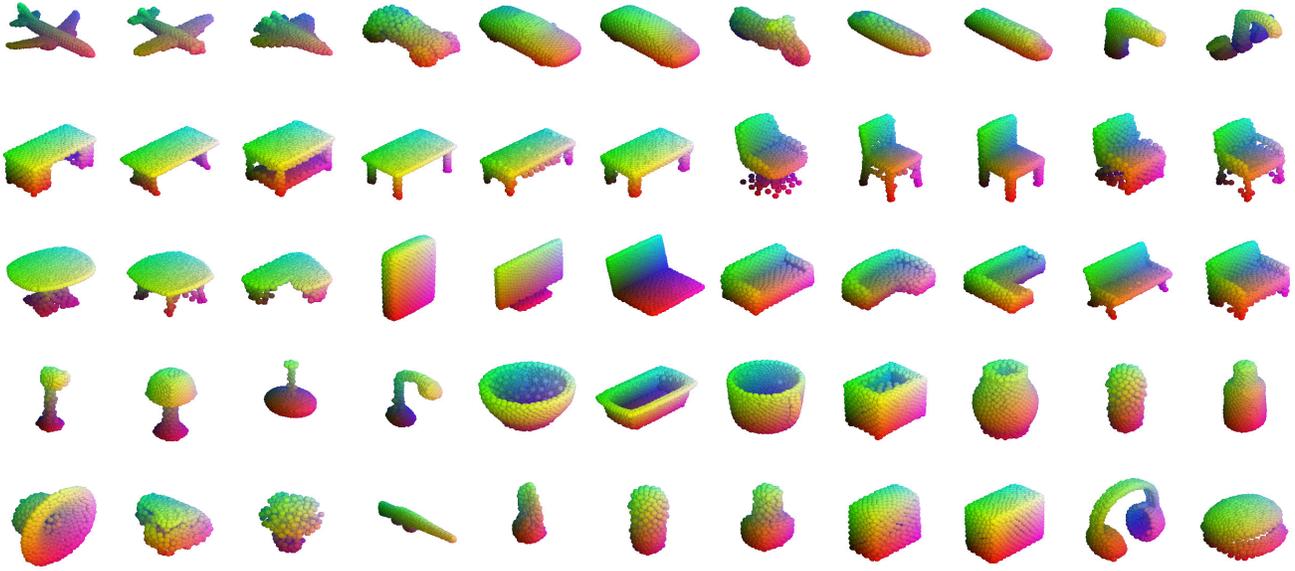

    \centering
    \setsepchar{ }
    \readlist\arg{10 42 43 41 2 8 54 15 17 38 30}
    \protocurriculumlonglast{Ours,\\full $D=5$}{last_epoch}{shapenet}{\arg}{0}
    \readlist\arg{1 35 4 24 49 51 5 19 31 44 52}
    \protocurriculumlonglast{Ours,\\full $D=5$}{last_epoch}{shapenet}{\arg}{0}
    \readlist\arg{45 37 48 12 25 50 47 36 6 21 40}
    \protocurriculumlonglast{Ours,\\full $D=5$}{last_epoch}{shapenet}{\arg}{0}
    \readlist\arg{3 14 33 0 29 28 16 20 18 22 32}
    \protocurriculumlonglast{Ours,\\full $D=5$}{last_epoch}{shapenet}{\arg}{0}
    \readlist\arg{11 9 13 23 7 22 27 34 46 39 26}
    \protocurriculumlonglast{Ours,\\full $D=5$}{last_epoch}{shapenet}{\arg}{0}
    \caption{\textbf{Modeling ShapeNet.}
    Prototypes from all $55$ linear shape models learned on ShapeNet~\cite{shapenet2015}, with our final $5$-dimensional model ``Ours, full $D=5$".
    In this figure, the prototypes have been manually rearranged with respect to their semantics. Note that some diverse classes such as tables or chairs are modeled by several models.
    }
    \label{fig:lsm_sn}
\end{figure*}

Our implementation uses PyTorch, Torch-Points3D \cite{tp3d}, and an efficient CUDA implementation of the Chamfer distance which significantly speeds up training. Code and data are available at: {\small\url{https://romainloiseau.github.io/deep-linear-shapes}}

\paragraph{Training strategy.} We use the Adam optimizer~\cite{kingma2014adam} with a learning rate of $0.001$, a batch size of 64, and neither weight decay nor data augmentation. Our model takes point clouds in $\bR^{1024\times3}$ as input for all experiments, except for the few-shot segmentation task that takes point clouds in $\bR^{2048\times3}$ as input.

\paragraph{Curriculum learning.}Inspired by the curriculum learning strategy of~\cite{monnier2020dticlustering}, we propose to learn our models by gradually increasing the models complexity. We first learn raw prototype models ($\cRk_\text{proto}$), an optimization which corresponds to performing a gradient-based K-means algorithm in the 3D space. Second, we augment each model with alignment awareness ($\cRk_\text{align}$). Finally, we gradually increase the linear families dimension up to the desired one, resulting in our final shape model ($\cRk_\text{full}$). Curriculum learning allows the model to choose the number of displacement fields $D$ according to the complexity of the studied dataset. Early stopping occurs when the benefit of adding a new degree of liberty (\ie increasing $D$ by one) does not meet a criterion on the loss or on a validation task, see Figure~\ref{fig:D}.

Alignment networks and basis vectors are initially set to identity and zero, respectively. When unfreezing a new module (alignment or a dimension of projection), the learning rate for the new weights is initially set to a tenth of the learning rate applied for the rest of the network, and gradually increased over $50$ epochs to the global learning rate. This ``warm-up" heuristic helps the network learn more smoothly from one step of the curriculum to the next.

\begin{figure}[h!]
    \centering
    \begin{tikzpicture}

    \pgfplotsset{width=7cm,compat=1.3}

	\begin{axis}[
	    axis y line*=left,
		xlabel=$D$,
		xtick = {0, 1, 2, 3, 4, 5},
		width=.91\linewidth,height=10em,
        ylabel=Accuracy,
		every y tick/.style={blue},
		every y tick label/.style={blue},
		]
		
	\addplot[color=blue!100!red,mark=x, style={thick}] coordinates {
	    (0.0, 0.745948147773743)
		(1.0, 0.755909371376038)
        (2.0, 0.762808728218079)
        (3.0, 0.767666876316071)
        (4.0, 0.766360485553741)
        (5.0, 0.76844254732132)
	}; \label{Accuracy}
	
	\end{axis}
	
	\begin{axis}[
	    axis y line*=right,
	    axis x line=none,
	    legend style={at={(.95,.75)},anchor=north east, font=\footnotesize},
		width=.91\linewidth,height=10em,
		ylabel=CD,
		every y tick/.style={red},
		every y tick label/.style={red},
		]
	\addlegendimage{/pgfplots/refstyle=Accuracy}\addlegendentry{Accuracy}
	\addplot[color=red!100!red,mark=x, style={thick}] coordinates {
	    (0.0, 0.009520857967436)
		(1.0, 0.008029004558921)
        (2.0, 0.00722084864974)
        (3.0, 0.006672278512269)
        (4.0, 0.006300711631775)
        (5.0,0.005988960061222)
	}; \addlegendentry{CD}
	
	\end{axis}
	
\end{tikzpicture}
    \vspace{-1em}
    \caption{\textbf{Influence of $D$ on ModelNet.} The reconstruction error (CD) decreases with added degrees of freedom. In contrast, the clustering Accuracy stops increasing when $D>=3$, hinting that we have reached a sufficient level of complexity.}
    \label{fig:D}
\end{figure}
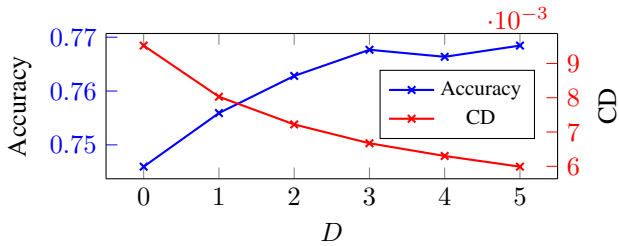

\paragraph{Initialization strategy.}As it is the case for many clustering algorithms, initialization can be critical. In our case, we initialize the prototype point clouds with samples of the training set chosen according to a k-means++ strategy~\cite{arthur2006k} with respect to the Chamfer distance.

\paragraph{Cluster reassignment.}To prevent empty clusters, we reassign at the end of each epoch any cluster that was selected fewer times than $20\%$ of the expected size of clusters ($N/K$) in the evenly distributed cluster assignment of Equation~1. Clusters are reassigned by selecting and duplicating another cluster. The duplicated cluster is chosen with a probability proportional to the mean of its reconstruction error over the last epoch. To break the symmetry, we add Gaussian noise with variance $10^{-4}$ to both its prototype and vector basis. The alignment and projection networks are copied without adding noise. We decrease the reassignment threshold tenfold after each curriculum step in order to preserve less populated but expressive clusters.

\begin{table}[h]
    \caption{Low-shot supervised segmentation results on ShapeNetPart. We report the IoU averaged over all classes.}
    \centering
    \small
    \begin{tabular}{@{}c|ccc@{}}
    \toprule
        Training &  SONet
        & 3D-PointCapsNet & Ours\\
        data &  \cite{li2018so}
        & \cite{zhao20193d}
        & full $D=5$\\\midrule
        $1\%$ & $64$ & $67$ & $68$ \\
        $5\%$ & $69$ & $70$ & $72$
    \\\bottomrule
    \end{tabular}
    \label{tab:seg_sup}
    \vspace{-1.5em}
\end{table}

\paragraph{Memory and Speed.}With $K=10$ prototypes and $D=5$, our model has $4.6M$ parameters. For comparison, the reconstruction models proposed by Wang \etal~\cite{wang2020few} and Groueix \etal~\cite{groueix2019unsupervised} have respectively $2.6M$ and $10.0M$ parameters. Our model can be trained on a single NVIDIA GeForce RTX 2080Ti within a few hours on the $3\,991$ samples of ModelNet10, and in less than a day on ShapeNetCore. Inference on all samples from ShapeNetCore ($\approx 50k$ shapes) takes less than $4$ minutes.

\paragraph{Choice of $K$.}
The number of models can be automatically selected through usual model selection heuristics such as the Bayesian Information Criterion (BIC), as we show in Figure~\ref{fig:K}.
Being entirely unsupervized, there is no restriction on how linear families relate to classes: complex classes can be represented by several models, and similar classes by a single family. However, as demonstrated in our clustering experiments, when the number of classes and models are the same, linear families and classes tend to be assigned on a one-to-one basis

\begin{figure}[h]
    \centering
    \begin{tikzpicture}

    \pgfplotsset{width=\linewidth,compat=1.3}

	\begin{axis}[
		ylabel=BIC,
		axis y line*=left,
		axis x line*=bottom,
		ytick =\empty,
		xtick = {0, 2, 4, 6, 8, 10, 12, 14, 16, 18, 20},
		legend pos=north east,
		width=\linewidth,height=10em,
		every x tick/.style={blue},
		every x tick label/.style={blue},
		]
		
	\addplot[color=blue,mark=x, style={thick}] coordinates {
		(1.0, 465.06229758117854)
        (2.0, 360.44714319835714)
        (3.0, 311.24183947049573)
        (4.0, 289.5619430939143)
        (5.0, 280.15327855111286)
        (6.0, 275.94306214997147)
        (7.0, 273.47218455327004)
        (8.0, 276.6683252441486)
        (9.0, 277.4365656260272)
        (10.0, 280.54171807172577)
        (11.0, 285.3843477998443)
        (12.0, 289.8975687679629)
        (13.0, 294.8790825006215)
        (14.0, 300.15745025646004)
        (15.0, 307.3550342815586)
        (16.0, 311.1026484458572)
        (17.0, 318.0048129729558)
        (18.0, 325.0136789937943)
        (19.0, 331.92623880697295)
	};
	\label{ModelNet10}
	\end{axis}
	
	\begin{axis}[
		axis x line*=top,
		axis y line*=right,
		ytick =\empty,
		xtick = {0, 10, 20, 30, 40, 50, 60, 70, 80, 90},
		legend pos=north east,
		legend style={at={(.38, .9)},anchor=north, font=\footnotesize},
		width=\linewidth,height=10em,
		every x tick/.style={red},
		every x tick label/.style={red},
		]
	
	\addlegendimage{/pgfplots/refstyle=ModelNet10}\addlegendentry{ModelNet10}
	\addplot[color=red,mark=x, style={thick}] coordinates {
	   (5,  507.969181329845)
	   (10, 411.398848119603)
	   (15, 383.304666061374)
       (20, 373.702579438891)
       (25, 371.384978919053)
       (30, 374.154840847347)
       (35, 380.024622224884)
       (40, 389.490111459179)
       (45, 397.517228186374)
       (50, 408.757744002792)
       (55, 419.561202217854)
       (60, 430.486958236041)
       (65, 442.421979962779)
       (70, 456.053164192931)
       (75, 470.45307934867 )
       (80, 482.083835022668)
       (85, 496.534628745412)
	};
	\addlegendentry{ShapeNet}
	
	\end{axis}
\end{tikzpicture}
    \caption{\textbf{Model Selection.} We can select the number of clusters $K$ using the BIC. We obtain $7$ clusters for ModelNet and $30$ for ShapeNet, which is consistent with the shapes' diversity.}
    \label{fig:K}
    \vspace{-1em}
\end{figure}
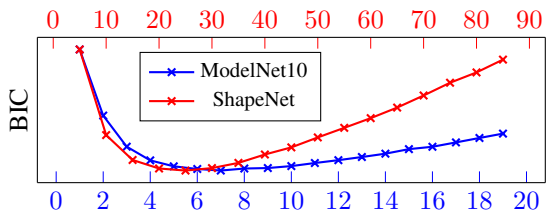

\section{Low-shot setting}

\paragraph{Low-Shot Semantic Segmentation.}
Our models can learn to perform semantic segmentation from a small number of annotated examples. We first initialize a set number of prototypes per class with random examples from the training set. This allows us to associate each prototype's point with a \emph{part} semantic label.
We then perform our standard training scheme, but with an altered Chamfer distance, which can only match points with the same part label from the true and reconstructed point clouds.
At inference time, we can associate a part label to each point of the input shape by taking the points' closest neighbor in their reconstructed shape.
This setting is supervised in the sense that we use the point labels explicitly during training.
As presented in \tabref{tab:seg_sup}, our model trained on only $1$ and $5\%$ of the annotated shapes yields an improvement of $+1$ and $+2$ average IoU points respectively, compared to 3D-Capsule~\cite{zhao20193d}. In contrast to this more complex model, our linear shape models remain viewable and interpretable.

\section{Self-supervised classification}

\paragraph{Self-supervised classification.}
To assess the capacity of our approach to extract relevant information from 3D models, we evaluate it in a standard self-supervised feature learning setup. We train our method on ShapeNetCore with $55$ 5-dimensional families, extract features from our results, and train and evaluate a linear SVM on the ModelNet10/40 train-test split following standard practices. We define three features that can be extracted from our model. A first type of features is defined as the soft-minimum of the distance between an input point and the reconstruction predicted by each linear shape model, with a temperature taken here as $100$ (``Distances"). These first features are complemented by concatenating the coordinates predicted by the projection networks (``Distances and coordinates"). Finally, we directly use the features from our point cloud encoder (``Embedding").

We present the results obtained with these different features in \tabref{tab:classification}, and compare to approaches specifically designed for self-supervised classification. While our method's performance is below most of these dedicated approaches, our results are still promising. Interestingly, we can see that adding shape coordinates to the distances significantly boosts the results, and even outperforms the latent embedding learned by the encoder, which outlines that our learned shape spaces are informative and meaningful.

\begin{table}[ht!]
    \caption{
    Results of the self-supervised classification task. We report the accuracy of a linear SVM trained on the training set of ModelNet using as input feature the reconstruction error to $55$ linear shape families of dimension $5$ augmented or not by the predicted coordinates, or the latent vector outputted by the point cloud encoder. In parenthesis, we report the name of the backbone network used (PointNet~\cite{qi2017pointnet}, PointNet++~\cite{qi2017pointnet++}, or VGG19~\cite{simonyan2014very}).}
    \centering
    \small
    \begin{tabular}{@{}lccc@{}}
    \toprule
          & $N_\text{feat.}$ &MN40 & MN10\\\midrule
        \textbf{Ours}, full $D=5$ (PointNet)\\
        \quad Distances&$55$ & $70.5$  & $86.2$ \\
        \quad Distances and coordinates & $330$ & $86.8$  & $90.9$ \\
        \quad Embedding & $512$& $86.2$   & $89.6$ \\\midrule
        3D-GAN~\cite{wu2016learning} &$7168$ & $83.3$  & $91.0$ \\
        VIP-GAN~\cite{han2019view} (VGG19)&$512$ & $90.2$ & $92.2$ \\
        FoldingNet~\cite{yang2018foldingnet} & $512$ & $88.4$ & $94.4$\\
        Latent-GAN~\cite{achlioptas2018learning} (PointNet)&$512$ & $84.5$ & $95.4$ \\
        Rec-Space~\cite{sauder2019self} (PointNet)  &$512$ & $87.3$ & $91.6$ \\
        Multi-Task~\cite{hassani2019unsupervised} & $512$ & $89.1$ & --- \\
        Label-Efficient~\cite{gadelha2020label} (PointNet++) & $512$ & $89.8$&---\\
    \bottomrule
    \end{tabular}
    \label{tab:classification}
\end{table}

\section{Learned Linear Shape Models}

In this section, we show qualitative examples of learned linear shape models. In \figref{fig:lsm_sn}, we represent all $55$ models learned on ShapeNet~\cite{shapenet2015}, with our final $5$-dimensional model ``Ours, full $D=5$". This illustrates how our model can be used to represent concisely a diverse and complex dataset such as ShapeNet without any supervision. In \figref{fig:lsm_abc}, we  show the $10$ models learned on our subset of ABC~\cite{koch2019abc} for the three steps of our curriculum strategy, illustrating the benefit of both alignment networks and linear families to learn such a diverse shape dataset. We also display the vector basis for all $5$ learned dimensions, representing the richness of each linear shape family.

Lastly, we represent in \figref{fig:lsm_mn10} the models learned on ModelNet with random rotations. We observe that when alignment networks are used, the obtained prototypes are similar to the ones obtained on the aligned version of the dataset. This shows that our approach can be used successfully on raw, un-aligned datasets.

\begin{figure*}[ht!]
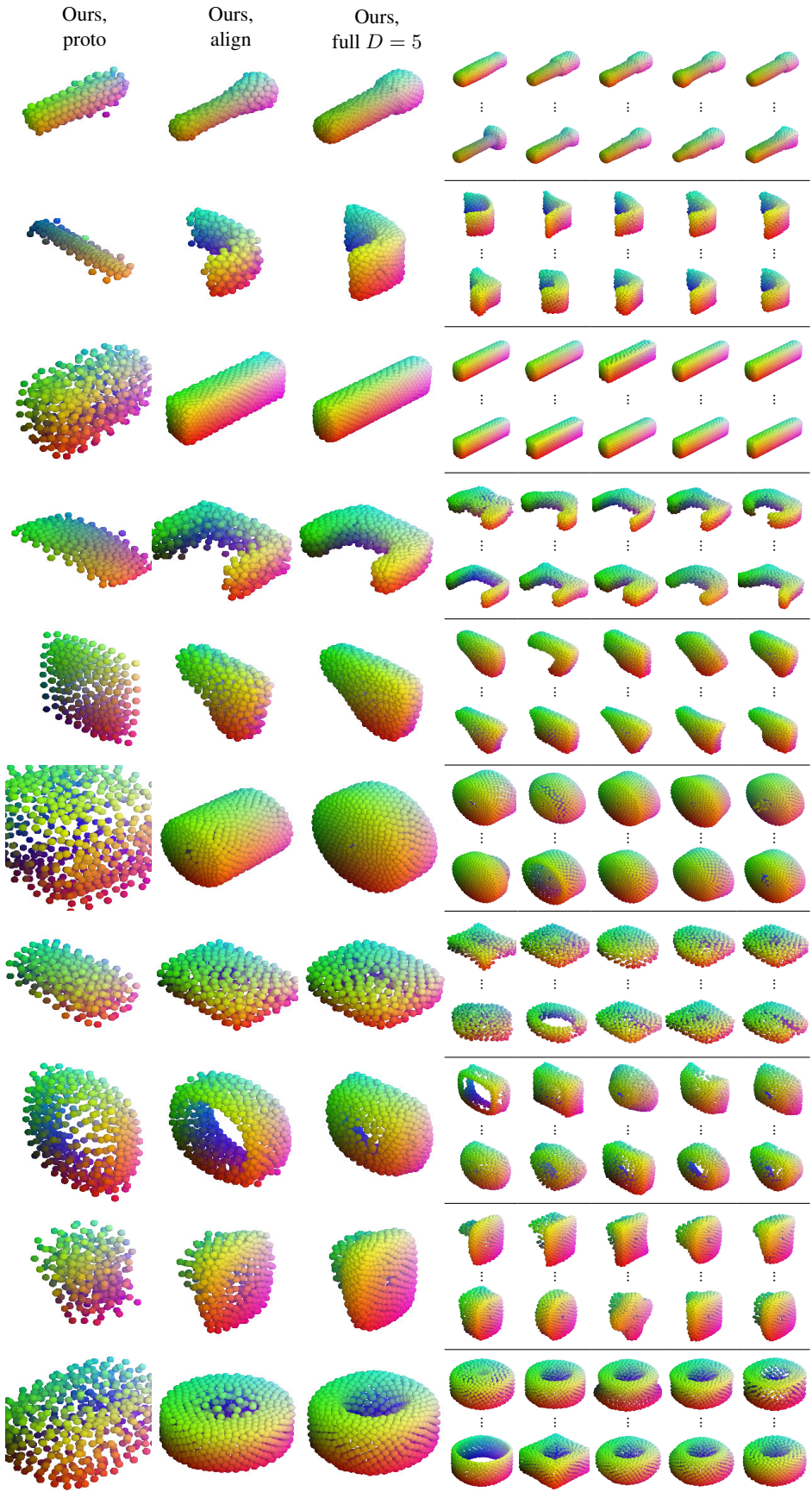

    \centering
    \begin{minipage}[c]{0.7\textwidth}
        \setsepchar{ }
        \readlist\arg{0 1 2 3 4 5 6 7 8 9}
        \protocurriculumandfields{abc}{\arg}
    \end{minipage}\hfill
    \begin{minipage}[c]{0.27\textwidth}
        \caption{\textbf{Modeling ABC.}
    $10$ prototype from linear shape models learned on the ABC dataset~\cite{koch2019abc}.
    Note that the prototypes are smoother and sharper when using alignment networks and $5$-dimensional linear families.
    On the right-most columns, we illustrate the $5$ dimensions of the linear family of each shape model. Each linear family spans a rich subset of the space of shapes.
    }
    \label{fig:lsm_abc}
    \end{minipage}
\end{figure*}

\begin{figure*}[ht!]
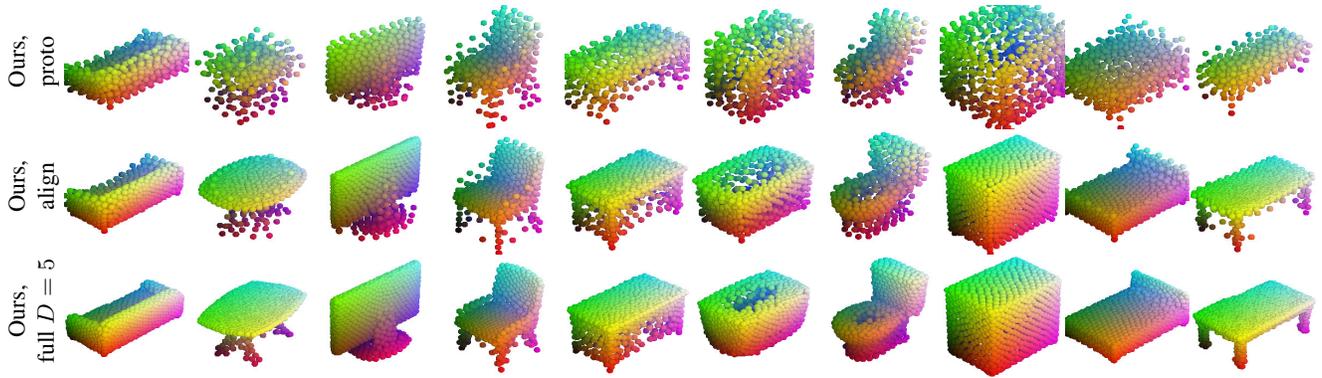
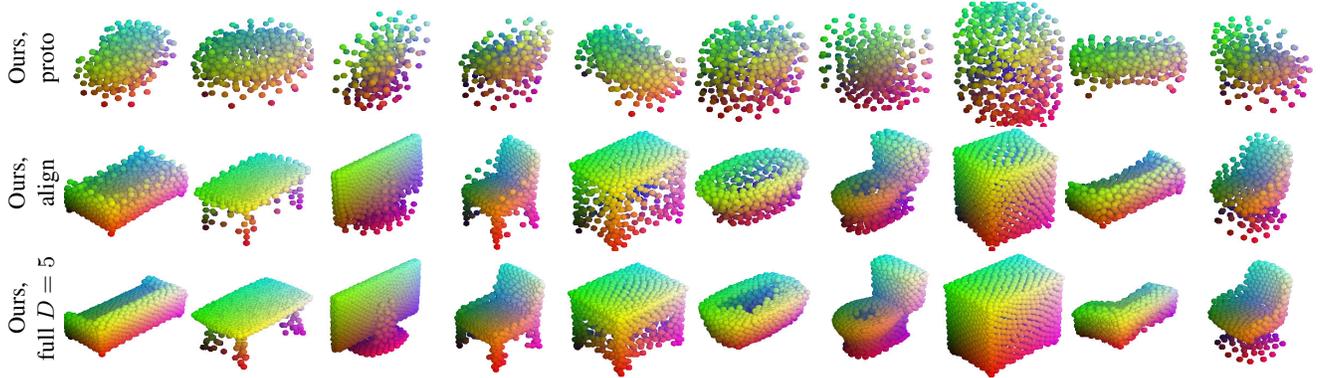

    \centering
    \begin{subfigure}[b]{\linewidth}
        \setsepchar{ }
        \readlist\arg{0 1 2 3 4 6 7 8 9 5}
        \protocurriculum{modelnet10}{\arg}
        \vspace{-1em}
        \caption{$10$ prototypes of the linear shape models learned from the \textbf{aligned ModelNet10} dataset.}
        \label{fig:modelnet:aligned}
    \end{subfigure}
    \begin{subfigure}[b]{\linewidth}
        \setsepchar{ }
        \readlist\arg{6 2 4 8 9 3 0 1 7 5}
        \protocurriculum{UNAXEDonAXEDmodelnet10}{\arg}
        \vspace{-1em}
        \caption{$10$ prototypes of the linear shape models learned from the \textbf{rotated ModelNet10} dataset.}
        \label{fig:modelnet:rotated}
    \end{subfigure}
    \caption{\textbf{Modeling ModelNet10.} 
    Prototype learned on ModelNet's~\cite{wu20153d} aligned version (\subref{fig:modelnet:aligned}) and with random $z$-axis rotations (\subref{fig:modelnet:rotated}). In this figure, the models are manually rearranged to be in correspondence across the two experiments.
    Note how our model without alignment networks (``Ours, proto") is unable to learn meaningful prototypes on un-aligned data. In contrast, our models with alignment networks learn sharp and informative prototypes despite the rotations. This shows that alignment networks allow our model to handle a raw, un-aligned dataset to produce a compact overview of its shape diversity. 
    }
    \label{fig:lsm_mn10}
\end{figure*}
\section{Reconstruction results}
We show some reconstruction results in \figref{fig:rec_abc} and \figref{fig:rec_sn} for ABC and ShapeNet respectively.
For each model, we represent some sample shapes for which the model provides the reconstruction with the lowest error. Viewing our approach in terms of clustering, this amounts to showing elements from the clusters associated with each model.
Note that in \figref{fig:rec_sn}, our linear models are associated with rich subsets of shapes which remain mostly semantically homogeneous.

\begin{figure*}[ht!]
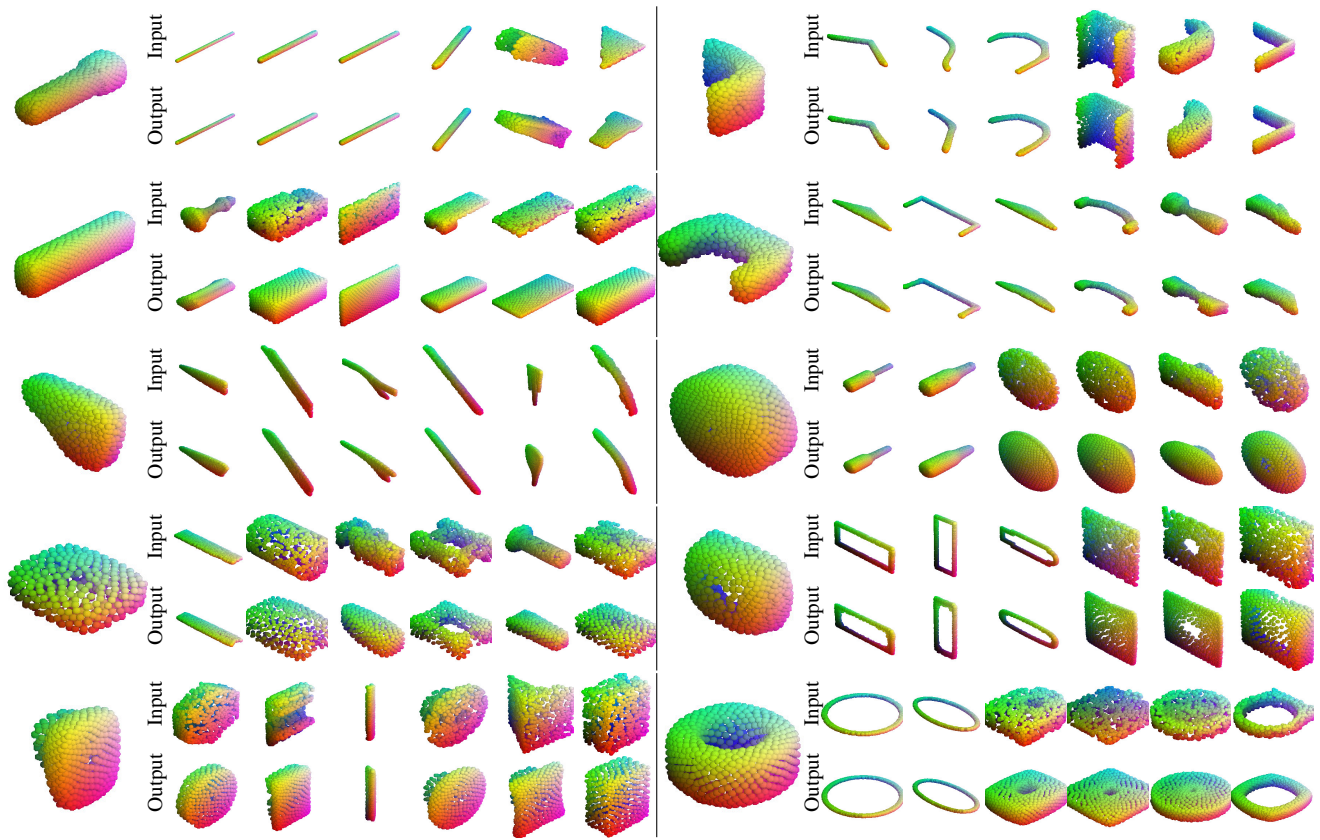

    \centering
    \setsepchar{ }
    \readlist\arga{0 5 6 10 14 16}
    \readlist\argb{1 2 3 6 8 17}
    \protorecsuppmat{abc}{0}{\arga}{1}{\argb}
    \readlist\arga{6 7 8 10 16 13}
    \readlist\argb{0 3 4 8 12 15}
    \protorecsuppmat{abc}{2}{\arga}{3}{\argb}
    \readlist\arga{0 1 2 3 4 5}
    \readlist\argb{0 1 5 10 13 14}
    \protorecsuppmat{abc}{4}{\arga}{5}{\argb}
    \readlist\arga{0 5 7 12 13 14}
    \readlist\argb{0 2 4 8 12 14}
    \protorecsuppmat{abc}{6}{\arga}{7}{\argb}
    \readlist\arga{6 7 8 9 11 13}
    \readlist\argb{0 1 6 8 9 13}
    \protorecsuppmat{abc}{8}{\arga}{9}{\argb}
    \caption{\textbf{Visualizing Reconstructions on ABC.}
    The left-most columns represent prototypes from all 10 linear models learned on the ABC dataset. For each prototype, we select $6$ samples for which this model gives the best reconstruction (``Input", top line). We then represent the associated reconstruction provided by the model (``Output", top line).
    Each family represents a wide variety of morphologically homogeneous shapes: round rings, square rings, bent archs, cylinders, etc...
    Looking at the prototypes gives us a concise overview of the shape diversity.
    }
    \label{fig:rec_abc}
\end{figure*}

\begin{figure*}[ht!]
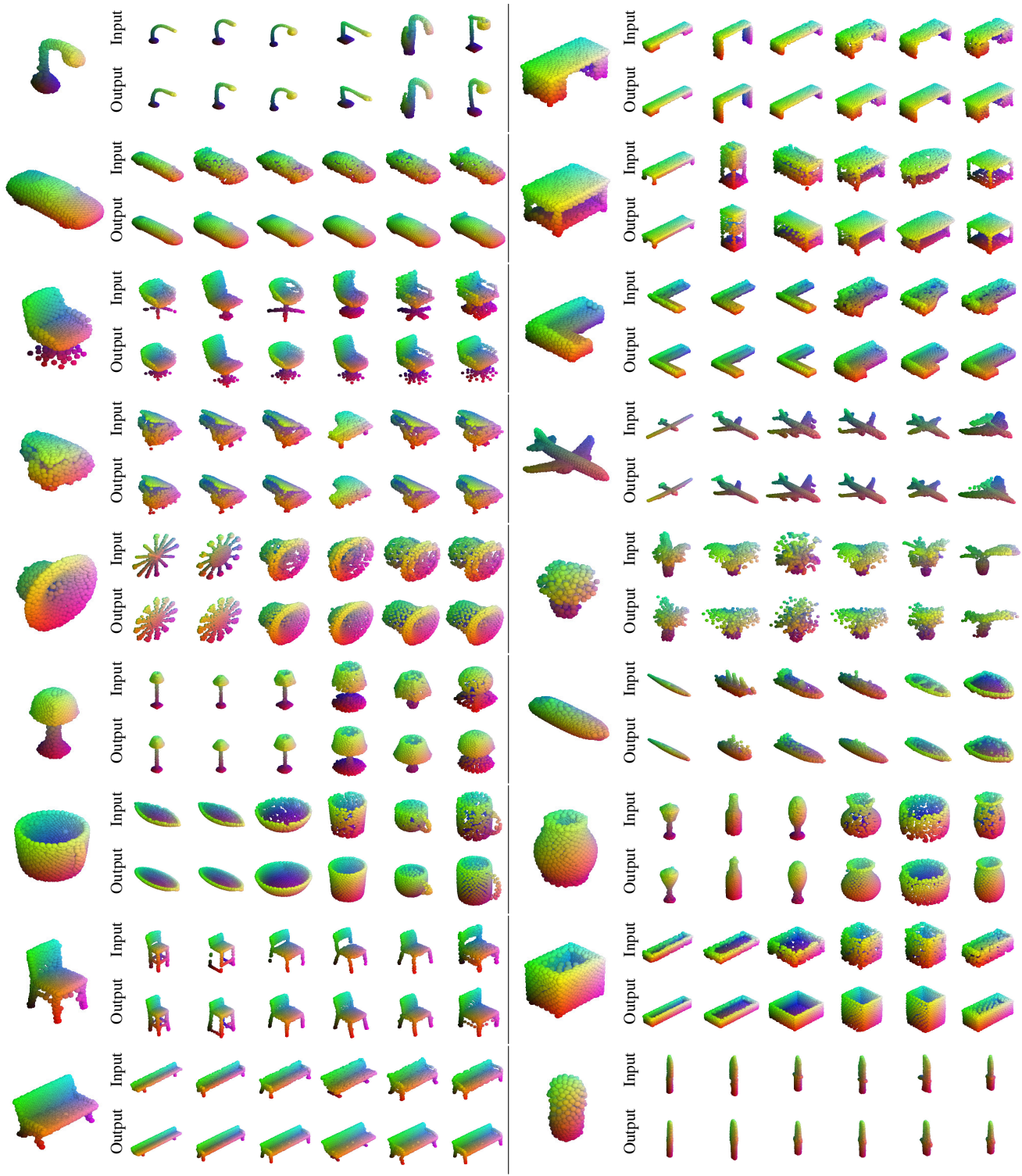

    \centering
    \setsepchar{ }
    \readlist\arga{0 1 2 3 5 16}
    \readlist\argb{0 1 2 5 15 16}
    \protorecsuppmat{shapenet}{0}{\arga}{1}{\argb}
    \readlist\arga{3 6 9 10 11 14}
    \readlist\argb{2 5 6 10 13 17}
    \protorecsuppmat{shapenet}{2}{\arga}{4}{\argb}
    \readlist\arga{0 1 4 6 8 15}
    \readlist\argb{0 1 3 7 9 14}
    \protorecsuppmat{shapenet}{5}{\arga}{6}{\argb}
    \readlist\arga{0 1 2 3 4 5}
    \readlist\argb{0 3 5 10 11 13}
    \protorecsuppmat{shapenet}{9}{\arga}{10}{\argb}
    \readlist\arga{0 1 10 11 12 14}
    \readlist\argb{2 7 9 10 13 14}
    \protorecsuppmat{shapenet}{11}{\arga}{13}{\argb}
    \readlist\arga{0 2 6 7 10 11}
    \readlist\argb{1 5 10 11 14 16}
    \protorecsuppmat{shapenet}{14}{\arga}{15}{\argb}
    \readlist\arga{2 3 5 6 7 8}
    \readlist\argb{1 2 4 5 6 9}
    \protorecsuppmat{shapenet}{16}{\arga}{18}{\argb}
    \readlist\arga{0 1 2 3 4 5}
    \readlist\argb{0 3 6 11 14 17}
    \protorecsuppmat{shapenet}{19}{\arga}{20}{\argb}
    \readlist\arga{0 1 4 6 7 13}
    \readlist\argb{0 2 4 5 9 10}
    \protorecsuppmat{shapenet}{21}{\arga}{22}{\argb}
    \caption{\textbf{Visualizing Reconstructions on ShapeNet.}
    The left-most columns represent prototypes from some of the $55$ linear models learned on the ShapeNet dataset. For each prototype, we select $6$ samples for which this model gives the best reconstruction (``Input", top line). We then represent the associated reconstruction provided by the model (``Output", top line).
    We observe that the samples associated with a given model are for the most part semantically homogeneous, and well represented by their prototype.}
    \label{fig:rec_sn}
\end{figure*}

}

\clearpage
{\small\balance{
\bibliographystyle{tpl_3dv/ieee_fullname}
\bibliography{main}}
}
\end{document}